\documentclass[envcountsame,oribibl]{llncs}
\usepackage{amssymb}
\usepackage{amsfonts}
\usepackage{amsmath}
\usepackage{dsfont} 

\usepackage{lipsum}

\usepackage{graphicx}
\makeatletter
\newcommand{\figcaption}{\def\@captype{figure}\caption}
\newcommand{\tabcaption}{\def\@captype{table}\caption}
\makeatother

\usepackage[tight,footnotesize]{subfigure}

\usepackage[vlined,titlenumbered,algoruled]{algorithm2e}

\usepackage{colortbl}
\usepackage{ctable}
\newcolumntype{+}{>{\global\let\currentrowstyle\relax}}
\newcolumntype{^}{>{\currentrowstyle}}
\newcommand{\rowstyle}[1]{\gdef\currentrowstyle{#1}#1\ignorespaces}
\newcommand{\PreserveBackslash}[1]{\let\temp=\\#1\let\\=\temp}
\newcolumntype{C}[1]{>{\PreserveBackslash\centering}p{#1}}
\newcolumntype{R}[1]{>{\PreserveBackslash\raggedleft}p{#1}}
\newcolumntype{L}[1]{>{\PreserveBackslash\raggedright}p{#1}}

\usepackage{footmisc}
\DefineFNsymbols{mystyle}{\textdagger}
\setfnsymbol{mystyle}

\usepackage{mathtools}
\newtagform{myform}{\normalsize(}{)}
\usetagform{myform}

\begin{document}
\title{Multiview Two-Task Recursive Attention Model for Left Atrium and Atrial Scars Segmentation}
\titlerunning{Simultaneous Fully Automated Segmentation of the LA and the atrial scars}

\author{Jun Chen*$^1$, Guang Yang*$^2$, Zhifan Gao$^3$, Hao Ni$^4$, Elsa Angelini$^5$, Raad Mohiaddin$^2$, Tom Wong$^2$,Yanping Zhang **$^1$, Xiuquan Du **$^1$,  Heye Zhang$^3$, Jennifer Keegan$^2$ and David Firmin$^2$}

\institute{ $^1$Anhui University, Hefei, China. \\ 
$^2$Cardiovascular Research Centre, Royal Brompton Hospital, London, SW3 6NP, UK and National Heart \& Lung Institute, Imperial College London, London, SW7 2AZ, UK \\
$^3$Shenzhen Institutes of Advanced Technology, Chinese Academy of Sciences, Shenzhen, China \\
$^4$Department of Mathematics, University College London, London, WC1E 6BT,  UK and Alan Turing Institute, London, NW1 2DB, UK \\
$^5$Faculty of Medicine, Department of Surgery \& Cancer, Imperial College London, London, SW7 2AZ, UK.
}

\maketitle
\vspace{-1em}
\newcommand\blfootnote[1]{%
\begingroup
\renewcommand\thefootnote{}\footnote{#1}%
\addtocounter{footnote}{-1}%
\endgroup
}

\blfootnote{* These authors contributed equally to this work.}
\blfootnote{** Corresponding author (zhangyp2@gmail.com and dxqllp@ahu.edu.cn).}

\begin{abstract}

Late Gadolinium Enhanced Cardiac MRI (LGE-CMRI) for detecting atrial scars in atrial fibrillation (AF) patients has recently emerged as a promising technique to stratify patients, guide ablation therapy and predict treatment success. Visualisation and quantification of scar tissues require a segmentation of both the left atrium (LA) and the high intensity scar regions from LGE-CMRI images. These two segmentation tasks are challenging due to the cancelling of healthy tissue signal, low signal-to-noise ratio and often limited image quality in these patients. Most approaches require manual supervision and/or a second \emph{bright-blood} MRI acquisition for anatomical segmentation. Segmenting both the LA anatomy and the scar tissues automatically from a single LGE-CMRI acquisition is highly in demand. In this study, we proposed a novel fully automated multiview two-task (MVTT) recursive attention model working directly on LGE-CMRI images that combines a sequential learning and a dilated residual learning to segment the LA (including attached pulmonary veins) and delineate the atrial scars simultaneously via an innovative attention model. Compared to other state-of-the-art methods, the proposed MVTT achieves compelling improvement, enabling to generate a patient-specific anatomical and atrial scar assessment model.

\end{abstract}

\section{Introduction}
\label{sec:introduction}

Late Gadolinium-Enhanced Cardiac MRI (LGE-CMRI) has been used to acquire data in patients with atrial fibrillation (AF) in order to detect native and post-ablation treatment scarring in the thin-walled left atrium (LA) \cite{peters2007detection}. This technique is based on the different wash-in and wash-out gadolinium contrast agent kinetics between healthy and scarred tissues \cite{yang2017fully}. The hyper-enhanced regions in the LGE-CMRI images reflect fibrosis and scar tissue while healthy atrial myocardium is `nulled' \cite{keegan2015dynamic}. This has shown promise for stratifying patients, guiding ablation therapy and predicting treatment success. Visualisation and quantification of atrial scar tissue requires a segmentation of the LA anatomy including attached pulmonary veins (PV) and a segmentation of the atrial scars.

Solving these two segmentation tasks is very challenging using LGE-CMRI images, where the nulling of signal from healthy tissue reduces the visibility of the LA boundaries. Moreover, in the AF patient population, prolonged scanning time, irregular breathing pattern and heart rate variability during the scan can result in poor image quality that can further complicate both segmentation tasks. Because of this, previous studies have segmented the LA and PV anatomy from an additional \emph{bright-blood} data acquisition, and have then registered the segmented LA and PV anatomy to the LGE-CMRI acquisition for visualisation and delineation of the atrial scars
\cite{karim2014method,tao2016fully,yang2017multi}. This approach is complicated by motion (bulk, respiratory or cardiac) between the two acquisitions and subsequent registration errors.

Recent deep learning based methods have been widely used for solving medical image segmentation. A convolutional neural networks (CNN) based approach has been proposed to segment the LA and PV from bright-blood images \cite{mortazi2017cardiacnet}, but not yet applied for LGE-CMRI images. For most previous studies, the LA and PV have been segmented manually although this is time-consuming, subjective and lacks reproducibility \cite{karim2013evaluation}. Based on the segmented LA and PV, and the derived LA, the atrial scars is then typically delineated using unsupervised learning based methods, e.g., thresholding and clustering, as described in this benchmarking paper \cite{karim2013evaluation}.

In this study, a novel fully automated multiview two-task (MVTT) recursive attention model is designed to segment the LA and PV anatomy and the atrial scars directly from the LGE-CMRI images, avoiding the need for an additional data acquisition for anatomical segmentation and subsequent registration. Our MVTT method consists of a sequential learning and a dilated residual learning to segment the LA and proximal PV while the atrial scars can be delineated simultaneously via an innovative attention model.

\section{Method}

\begin{figure}[!htbp]
\centering
\setlength{\abovecaptionskip}{0cm} 
\setlength{\belowcaptionskip}{-1cm}
\scalebox{.98}{
\includegraphics[width=1\textwidth]{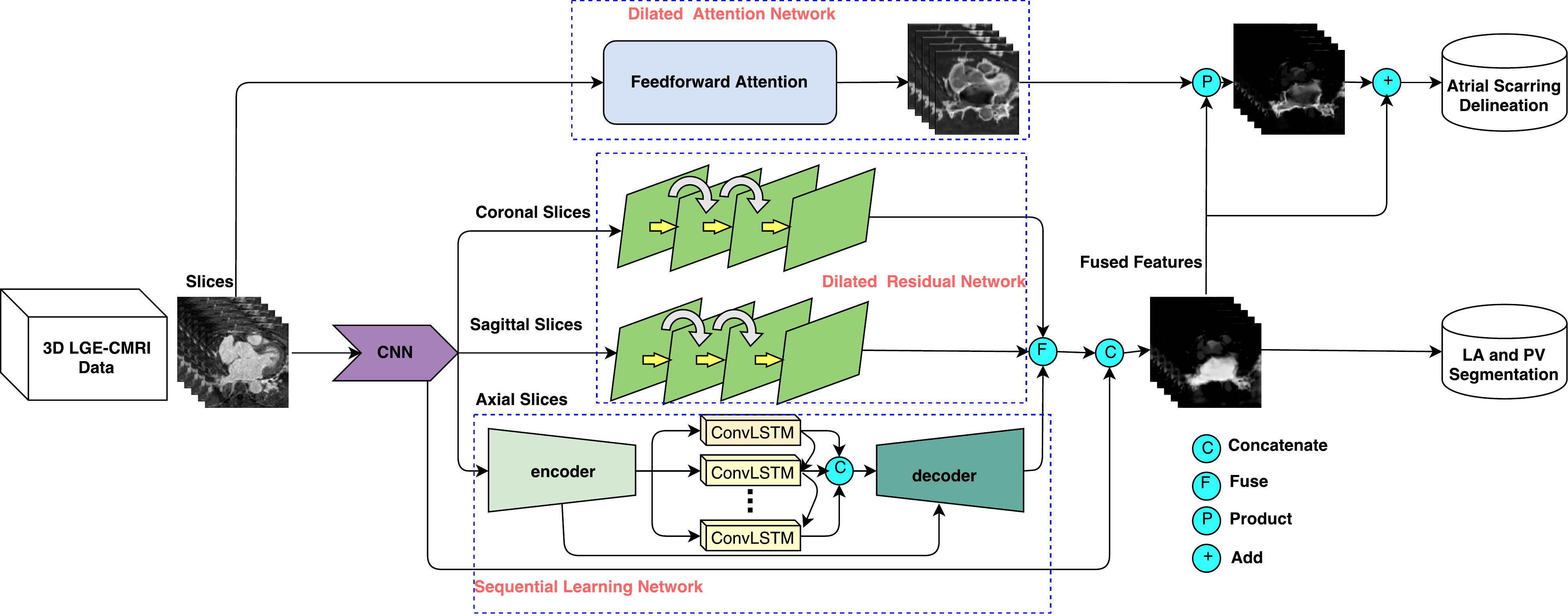}
}
\figcaption{\scriptsize{Workflow of our proposed MVTT recursive attention model (Detailed architecture of each subnetworks can be found in the Supplementary Materials).}}
\label{fig:workflow}
\end{figure}

The workflow of our MVTT recursive attention model is summarised as shown in Figure \ref{fig:workflow}. It  performs the segmentations for the LA and PV anatomy and atrial scars simultaneously.

\noindent\textbf{LA and PV Segmentation via a Multiview Learning.}
Our 3D LGE-MRI data were acquired and reconstructed into a volume with 60--68 2D axial slices with a spatial resolution of (0.7--0.75)$\times$(0.7--0.75)$\times$2mm$^3$. In this study, we propose a multiview based method to delineate LA and PV that mimics the inspection procedure of radiologists, who view the images by stepping through the 2D axial slices to obtain the correlated information in the axial view (with finer spatial resolution) while also using complementary information from sagittal and coronal views (with lower spatial resolution). We modelled the information extracted from the axial view by a \emph{sequential learning} and for the sagittal and coronal views we designed a \emph{dilated residual learning}.

\vspace{-0.5cm}
\vspace{0.4cm}

Firstly, a $3\times3$ convolutional layer with 12 kernels is used to extract the high resolution features. We stack the obtained feature maps to 3D maps to slice them into  axial,sagittal and coronal slices respectively, which are used to perform  multiview learning. Our \emph{sequential learning network} consists of six convolutional layers for the encoder and decoder path, respectively. In the encoder path, each convolutional layer is followed by a rectified linear unit (ReLU) layer and a local response normalisation (LRN) layer to normalise the feature maps. In addition, three max-pooling layers are used to reduce the dimension of the feature maps. In the decoder path, three up-sampling layers are implemented via bilinear interpolation to recover the original image size, and the decoder is also incorporated convolutional and LRN layers. Each convolutional layer contains 12 kernels with size of 3$\times$3 pixels. Furthermore, convolutional long-short term memory (ConvLSTM) \cite{Shi2015convolutional} layers are embedded into the encoder-decoder network to account for inter-slices correlations. The ConvLSTM is a special recursive neural network architecture that can be defined mathematically as
\begin{small}
\begin{equation}
i_{t}=\sigma(W_{xi}\ast x_{t}+W_{hi}\ast h_{t-1}+W_{ct}\circ c_{t-1}+b_{i}),
\end{equation}
\begin{equation}
f_{t}=\sigma(W_{xf}\ast x_{t}+W_{hf}\ast h_{t-1}+W_{cf}\circ c_{t-1}+b_{f}),
\end{equation}
\begin{equation}
c_{t}=f_{t}\circ c_{t-1} + i_{t}\circ \mathrm{ReLU}(W_{xc}\ast x_{t}+W_{hc}\ast h_{t-1}+b_{c}),
\end{equation}
\begin{equation}
o_{t}=\sigma(W_{xo}\ast x_{t}+W_{ho}\ast h_{t-1}+W_{cfo}\circ c_{t}+b_{o}),
\end{equation}
\begin{equation}
h_{t}=o_{t}\circ \mathrm{ReLU}(c_{t})
\end{equation}
\end{small}
where `$\ast$' represents convolutional operator and `$\circ$' denotes the Hadamard product, $W$ terms denote weight matrices, $b$ terms denote bias vectors, $\sigma$ represents a sigmoid function and ReLU is used in our study instead of tanh. The ConvLSTM uses three gates including the input gate $i_t$, the forget gate $f_t$  and the output gate $o_t$, and memory cell $c_t$ represents an accumulator of the state information and $h_t$ denotes the hidden states.

Secondly, in order to learn the complementary information from the sagittal and coronal views, we propose to use a dilated residual network. In the network, we adopt the \emph{dilated convolution} \cite{yu2015multi} and remove the max-pooling layers to avoid loss of useful information during the pooling operation. The network consists of four $3 \times 3$  dilated convolutional layers  based on residual architecture \cite{he2016deep}, and each has 12 kernels and is followed by a ReLU layer and a LRN layer.  By using the dilated convolution,  the size of the feature maps is remained.

Finally, two 3D volumes are created to store the learned feature maps from the sagittal and coronal view, respectively. Then we slice them into multiple 2D axial slices, and concatenate them with the feature maps derived from the sequential learning  at their corresponding channels. Next, a convolutional operation is applied to these concatenated feature maps to get the fused multiview features. Meanwhile, high resolution features from the previous layer are combined with the fused multiview features for localizing LA/PV and atrial scars.  At the end, three convolutional layers perform the LA/PV segmentation . Two of them contain 24 kernels with the size of 3$\times$3 and each is followed by a ReLU layer and a LRN layer. At the last layer, a 3$\times$3 convolution is used to map each pixel to the desired segments, and the sigmoid activation function is used.

\noindent\textbf{Atrial Scars Segmentation Via an Attention Model.}
The regions of atrial scars are relatively small and discrete; therefore, in this study we tackle the delineation of atrial scars using the attention mechanism to force the model to focus on the locations of the atrial scars, and enhance the representations of the atrial scars at those locations. Moreover, conventional pooling operations can easily lose the information of these small atrial scars regions. Therefore, a novel dilated attention network is designed to integrate a \emph{feedforward attention structure}\cite{Wang2017Residual} with the dilated convolution to preserve the fine information of the atrial scars. In our dilated attention network, the attention is provided by a mask branch, which is changing adaptively according to the fused multiview features. There are four convolutional layers for the mask branch and each of the first three layers is followed by a ReLU layer and a LRN layer. Finally, according to \cite{Wang2017Residual}, we utilise a sigmoid layer which connects to a $1\times 1$ convolutional layer to normalise the output into a range of [0,1] for each channel and spatial position to get the attention mask . This sigmoid layer can be defined as following
\begin{small}
\begin{equation}
AM(x_{i,c})=\frac{1}{1+e^{(-x_{i,c})}},
\end{equation}
\end{small}
where $i$ ranges over all spatial positions and $c$ ranges over all channels.

Because the soft attention mask can potentially affect the performance of the multiview learning, a residual architecture is also applied to mitigate such influence. The output $O$ of the attention model can be denoted as
\begin{small}
\begin{equation}
O(x_{i,c})=(1+AM(x_{i,c}))\cdot F(x_{i,c}),
\end{equation}
\end{small}
in which $i$ ranges over all spatial positions, $c$ ranges over all the channels, $AM(x_{i,c})$ is the attention mask, which ranges from [0,1], $F(x_{i,c})$ represents the fused multiview features, and $\cdot$ denotes the dot product. Finally, based on generated attention maps, three convolutional layers are connected at the end to perform the atrial scars delineation, which are similar to the ones used for the LA and PV segmentation.

\noindent\textbf{Implementation Details.}
For the implementation, we used the Adam method to perform the optimization with a mean squared error based loss function and decayed learning rate (the initial learning rate was 0.001 and dropped to 0.000445 at the end). Our deep learning model was implemented using Tensorflow 1.2.1 on a Ubuntu 16.04 machine, and was trained and tested on an NVidia Tesla P100 GPU (3584 cores and 16GB GPU memory).

\section{Experimental Results and Discussion}

We retrospectively studied 100 3D LGE-CMRI scans acquired in patients with longstanding persistent AF. Both pre- and post-ablation acquisitions were included (detailed LGE-CMRI scanning sequence and patient cohort information can be found in the Supplementary Materials). Manual segmentations of the LA and PV anatomy and atrial scars by an experienced physicist were used as the ground truth for training and evaluation of our MVTT recursive attention model. All patient recruitment and data acquisition were approved by the institutional review board in accordance with local ethics procedures. Ten-fold cross-validation was applied to provide an unbiased estimation of the accuracy, sensitivity, specificity and Dice score of the two segmentations. For comparison studies, we also evaluated the performance of state-of-the-art methods for LA and PV segmentation (using atlas based whole heart segmentation, WHS \cite{zhuang2016multi}) and the atrial scars delineation (using unsupervised learning based methods \cite{karim2013evaluation} and a re-implementation of the U-Net \cite{ronneberger2015unet}).

\noindent\textbf{LA and PV Segmentation.} The experimental results show that our MVTT framework can accurately segment the LA and PV (Table \ref{table:Tab1} and Figure \ref{fig:atrium_visualisation}). The obtained accuracy, sensitivity, specificity and Dice scores are 98.51\%, 90.49\%, 99.23\% and 90.83\%. Figure \ref{fig:atrium_visualisation} shows example segmentation results of the LA and PV for a pre-ablation case and a post-ablation case. Both the segmentation result obtained by our MVTT framework (green contours) and the ground truth (red contours) are overlaid on LGE-CMRI images, and our fully automated segmentation has demonstrated high consistency compared to the manual delineated ground truth.

\noindent\textbf{Atrial Scars Segmentation.} Our MVTT framework also shows great performance for segmenting the atrial scars (Table \ref{table:Tab1} and Figure \ref{fig:Fig2}). We achieve an overall segmentation accuracy of 99.90\%, with a sensitivity of 76.12\% , a specificity of 99.97\% and a Dice score of 77.64\% (median 83.59\% and 79.37\% for post- and pre-ablation scans). Segmentation results in Figure \ref{fig:Fig2} (c) and (k) show a great agreement compared to the ground truth. In addition, correlation analysis of the calculated scar percentage between our MVTT and the ground truth as shown in Figure \ref{fig:Fig3} (c). The derived correlation coefficient $r=0.7401 \in (0.6, 0.8)$ represents a strong correlation between our fully automated segmentation and the manual delineated ground truth. Furthermore, the Bland-Altman plot of the calculated scar percentage (Figure \ref{fig:Fig3} (d)) shows the 94\% measurements are in the 95\% limits of agreement, which also corroborates the accurate scar percentage measure of our MVTT framework.

\vspace{-0.5cm}
\begin{table}[!htbp]
  \setlength{\abovecaptionskip}{0pt} 
  \setlength{\belowcaptionskip}{0pt} 
  \caption{\scriptsize{Quantitative results of the cross-validated LA and PV segmentation and the atrial scars delineation. For the LA and PV segmentation, we compared with the WHS \cite{zhuang2016multi}, and for the atrial scars delineation we compared with the SD, k-means, Fuzzy c-means \cite{karim2013evaluation} and the U-Net \cite{ronneberger2015unet}.}}
  \centering
  \scalebox{.65}{
  \setlength{\floatsep}{10pt plus 3pt minus 2pt} 
\begin{tabular}{p{5.2cm}^p{3.2cm}^p{3.2cm}^p{3.2cm}^p{3.2cm}}
\addlinespace
\toprule\rowstyle{\bfseries}
            			& Accuracy      	& Sensitivity           	& Specificity               & Dice Score  \\ \midrule \midrule
WHS 					& $0.9978\pm{0.0009}$ & $0.8587\pm{0.0415}$ & $0.9993\pm{0.0006}$ & $0.8905\pm{0.0317}$ \\ \midrule
Multi-view 	        	& $0.9773\pm{0.0127}$ & $0.8163\pm{0.1355}$ 	& $0.9924\pm{0.0038}$ 		& $0.8502\pm{0.1033}$ 	\\ \midrule
Axial view + ConvLSTM	& $0.9778\pm{0.0088}$ &  $0.8370\pm{0.0802}$    &  $0.9909\pm{0.0036}$		&  $0.8609\pm{0.0510}$ 	\\ \midrule
S-LA/PV                	& $0.9845\pm{0.0081}$ &  $0.8901\pm{0.1012}$    &  $0.9930\pm{0.0035}$		&  $0.8999\pm{0.0857}$ 	\\ \midrule
MVTT            		& $0.9851 \pm{0.0052}$ &$0.9049 \pm{0.0487}$    &$0.9923 \pm{0.0041}$		& $0.9083 \pm{0.0309} $ \\ \midrule \midrule
2-SD					& $0.9984 \pm{0.0007}$ & $0.5137 \pm{0.2497}$ 	& $0.9994 \pm{0.0006}$    &$0.5277 \pm{0.1916}$ \\ \midrule
K-means					&$0.9975\pm{0.0009}$ &$0.7777\pm{0.1508}$      &$0.9981\pm{0.0010}$       &$0.5409\pm{0.1539}$ \\ \midrule
Fuzzy c-means			& $0.9974\pm{0.0010}$ &$0.7968\pm{0.1539}$		& $0.9979\pm{0.0010}$ 	  & $0.5350\pm{0.1601}$		\\ \midrule
U-Net					& $0.9987\pm{0.0008}$ & $0.8342\pm{0.1720}$	    & $0.9992\pm{0.0003}$  	 & $0.7372\pm{0.1326}$ \\ \midrule
Multi-view+ConvLSTM	&  $0.9990\pm{0.0008}$  &  $0.73267\pm{0.1705}$	&  $0.9997\pm{0.0002}$ 	& $0.7566\pm{0.1396}$ \\ \midrule
Multi-view+attention	& $0.9987\pm{0.0008}$  & $0.7928 \pm{0.1759}$ 	& $0.9993 \pm{0.0002}$ 		& $0.7275 \pm{0.1342}$ \\ \midrule
Axial view+ConvLSTM+Attention	& $0.9987\pm{0.0008}$ & $0.7861\pm{0.1719}$ 		& $0.9992\pm{0.0003}$ 	& $0.7090\pm{0.1415}$ \\ \midrule
S-Scar                	& $0.9989\pm{0.0009}$ &  $0.7464\pm{0.1675}$    &  $0.9995\pm{0.0003}$		&  $0.7441\pm{0.1448}$ 	\\ \midrule
MVTT 	            	& $0.9990 \pm{0.0009}$ &$0.7612 \pm{0.1708}$    &$0.9997 \pm{0.0002}$						& $0.7764 \pm{0.1459} $	\\ \bottomrule
\end{tabular}
  }
  \label{table:Tab1}
\end{table}
\vspace{-0.5cm}

\vspace{-0.5cm}
\begin{figure}[!thbp]
\centering
\setlength{\abovecaptionskip}{0cm} 
\setlength{\belowcaptionskip}{0cm}
\scalebox{1.0}{
\includegraphics[width=1\textwidth]{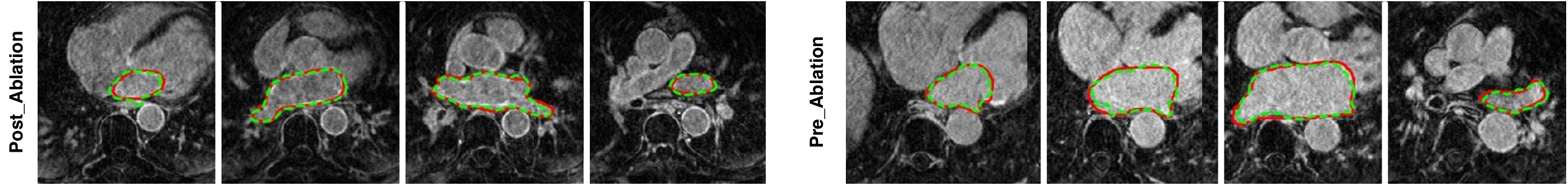}
}
\figcaption{\scriptsize{Qualitative visualisation of the LA and PV segmentation.}}
\label{fig:atrium_visualisation}
\end{figure}
\vspace{-0.5cm}

\noindent\textbf{Ablation Study I: Multi-View Sequential Learning.} To demonstrate the add-on value of our multi-view sequential learning (multi-view+ConvLSTM) for extracting better anatomical features. We replaced the multi-view with axial view only in our MVTT framework (axial view+ConvLSTM). Results in Table \ref{table:Tab1} show that multi-view sequential learning achieved better performance for the LA and PV segmentation.

\noindent\textbf{Ablation Study II: Attention Mechanism.} We introduced an attention mechanism to enforce our model to pay more \emph{attention} to the small atrial scar regions, and enhance their representations. To evaluate the performance of this attention mechanism, we removed the attention architecture from the MVTT framework, and only use the multi-view and ConvLSTM parts to predict the atrial scars. As shown in Table \ref{table:Tab1}, the MVTT trained with attention architecture outperforms the multi-view+ConvLSTM, which proves the effectiveness of our MVTT framework with the attention mechanism. 


\noindent\textbf{Comparison Studies.} Table \ref{table:Tab1} tabulates the quantitative comparison results for both the LA and PV segmentation and the atrial scars delineation. Compared to the WHS, our MVTT framework obtained much higher sensitivity (0.905 vs. 0.859) and similar specificity and therefore a higher Dice score. It is of note that the WHS method derived the LA and PV anatomy from additionally acquired \emph{bright-blood} images and that was then registered to the LGE-MRI to derive the scar segmentation. Our MVTT method derived both LA and PV anatomy and scar segmentations from a single 3D LGE-CMRI dataset, which is a more challenging task but one which eliminates the need for an additional acquisition and subsequent registration errors. For the atrial scars delineation, all the unsupervised learning methods, e.g., standard deviation (SD) based thresholding and clustering, obtained high specificities, but very low sensitivities and poor Dice scores. Qualitative visualisation in Figure \ref{fig:Fig2} shows that the SD method clearly underestimated the atrial scars and both k-means and Fuzzy c-means (FCM) over-estimated the enhanced scar regions. The U-Net based method improved the delineation, but was still struggling to segment the atrial scars accurately. In addition, the experiments with the same architecture but separated two tasks(S-LA/PV,S-scar) illustrated that our simultaneous method showed better results because the two tasks constrain each other.

\vspace{-0.5cm}
\begin{figure}[!htbp]
  \centering
  \setlength{\abovecaptionskip}{0pt} 
  \setlength{\belowcaptionskip}{0pt}
  \scalebox{.98}{
  \includegraphics[width=1\textwidth]{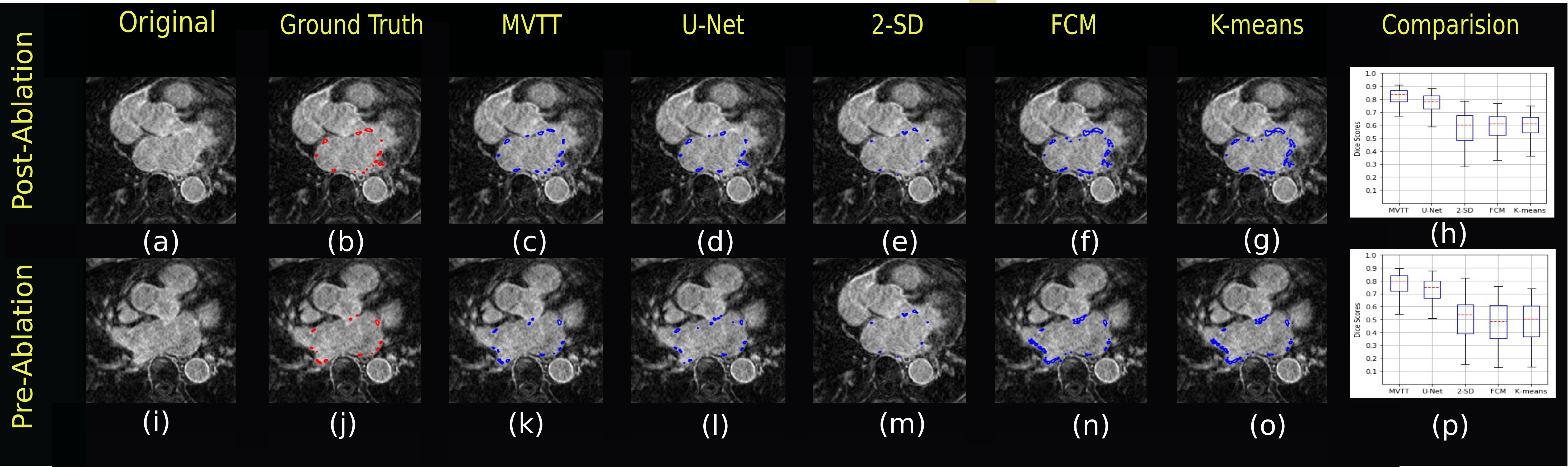}
  }
  \figcaption{\scriptsize{Qualitative visualisation of the atrial scars delineation.}}
  \label{fig:Fig2}
  \end{figure}
  \vspace{-0.5cm}

\noindent\textbf{Limitations.} One possible limitation is that our MVTT framework performed less well in some pre-ablation cases that is mainly due to very rare amount of native scar in these AF patients (see the outliers in Figure \ref{fig:Fig3} (c) and (d)). The performance of our proposed MVTT recursive attention model did not rely on a comprehensive network parameters tuning and currently used network parameters defined by test and trials may cause possible overfitting of the trained models (see convergence analysis in Figure \ref{fig:Fig3} (a-b)); however, this could be mitigated via techniques such as early stopping. 

 \vspace{-0.5cm}
  \begin{figure}[!htbp]
  \centering
  \setlength{\abovecaptionskip}{0pt} 
  \setlength{\belowcaptionskip}{0pt}
  \scalebox{.98}{
  \subfigure[]{\includegraphics[width=0.25\textwidth]{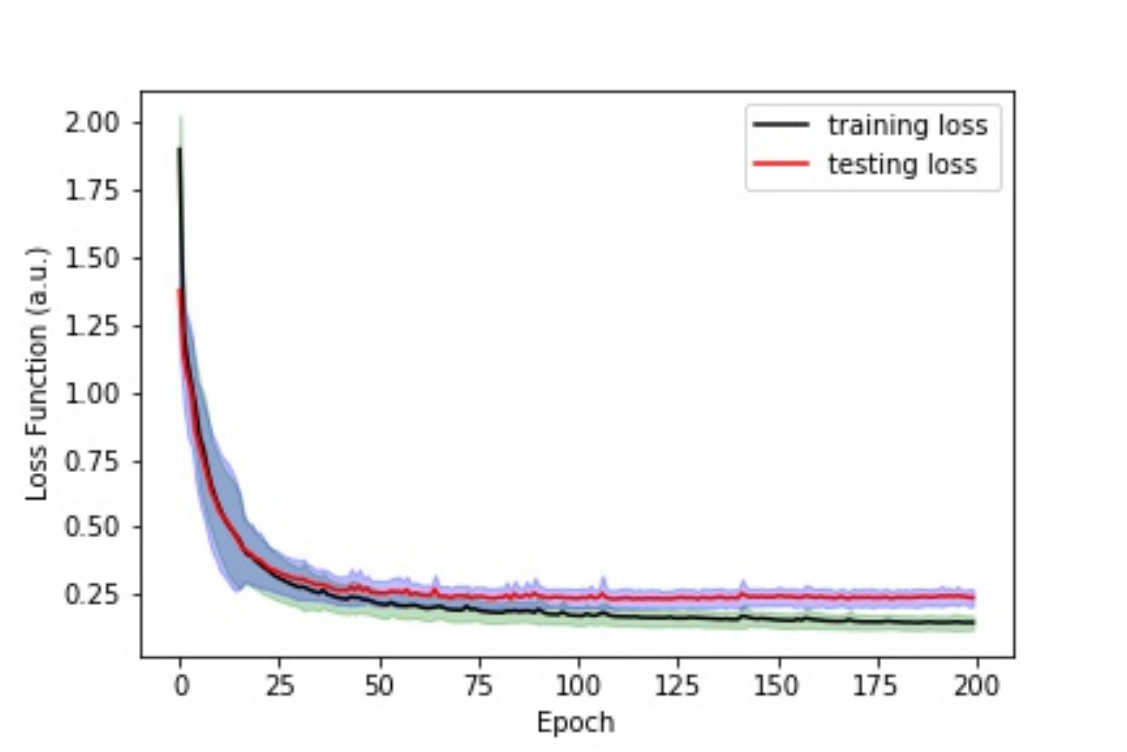}}
  \subfigure[]{\includegraphics[width=0.25\textwidth]{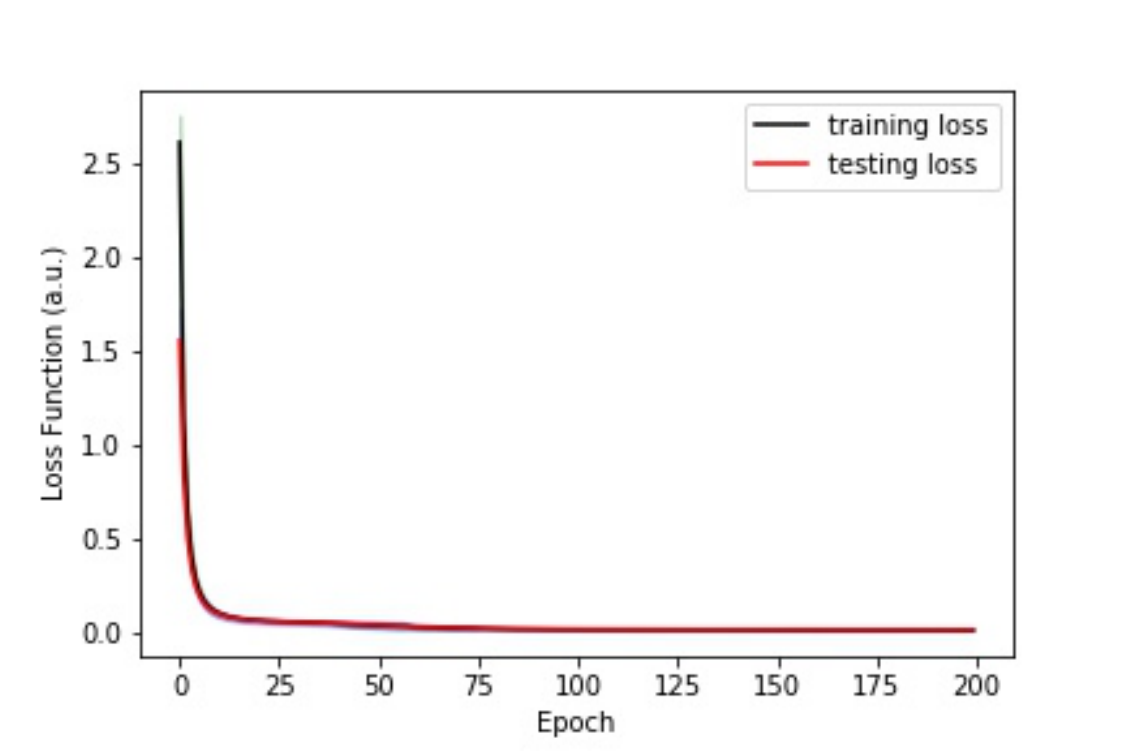}}
  \subfigure[]{\includegraphics[width=0.25\textwidth]{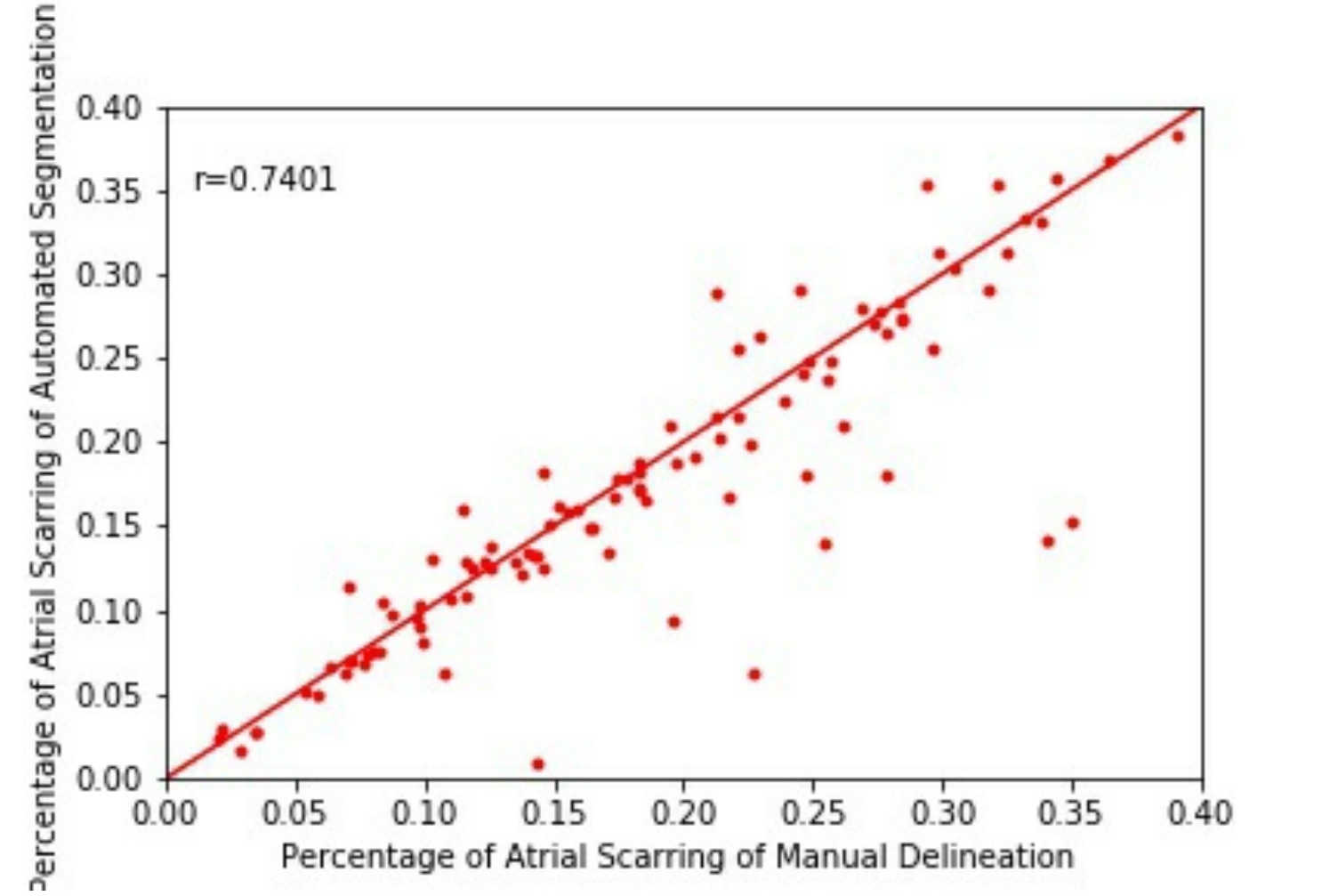}}
  \subfigure[]{\includegraphics[width=0.25\textwidth]{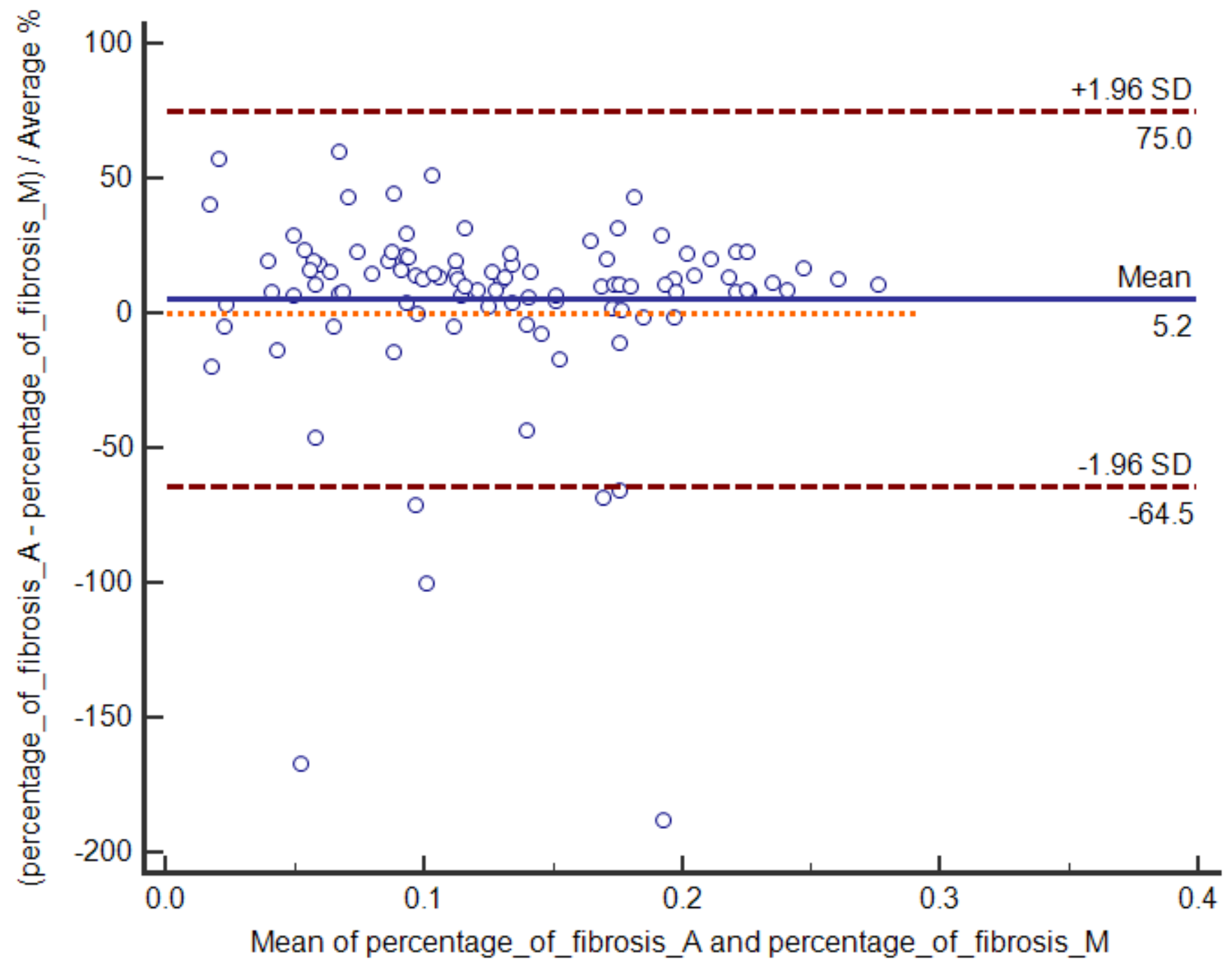}}
  }
  \figcaption{\scriptsize{(a) Training/testing convergence for the LA and PV segmentation, (b) Training/testing convergence for the atrial scars segmentation, (c) Correlation of the atrial scars percentage, (d)Bland-Altman plot of the measurements of atrial scars percentage.}}
  \label{fig:Fig3}
  \end{figure}
\vspace{-0.5cm}

\section{Conclusion}

In this work, we propose a fully automatic MVTT to segment both the LA and PV and atrial scars simultaneously from LGE-CMRI images directly. By combining the sequential learning and dilated residual learning for extracting multiview features, our attention model can delineate atrial scars accurately while segmenting the LA and PV anatomy. Validation of our framework has been performed against manually delineated ground truth. Compared to other state-of-the-art methods, our MVTT framework has demonstrated superior performance when using only the LGE-CMRI data. In conclusion, the proposed MVTT framework makes it possible to create a robust patient-specific anatomical model for the LA and PV that is accredited by our efficient and objective segmentation. It has further enabled a fast, reproducible and reliable atrial scarring assessment for individual AF patients.



\end{document}